\documentclass[10pt, a4paper,twocolumn]{article}
\usepackage{ltc23}
\usepackage{graphicx}
\usepackage{url}

\title{Language-Independent
Sentiment Labelling with
Distant Supervision:\\A Case Study for English,
Sepedi and Setswana}

\name{Koena Ronny Mabokela$^{\ast}$, Tim Schlippe$^{\textdollar}$, Mpho Raborife$^{\ast}$, Turgay Celik$^\|$} 
\address{ $^{\ast}$University of Johannesburg, South Africa\\
                  $^{\textdollar}$IU International University of Applied Sciences, Germany\\ 
                  $^{\|}$University of the Witwatersrand, South Africa\\ krmabokela@gmail.com}

\abstract{Sentiment analysis is a helpful task to automatically analyse opinions and emotions on various topics in areas such as \textit{AI for Social Good}, \textit{AI in Education} or marketing. While many of the sentiment analysis systems are developed for English, many African languages are classified as low-resource languages due to the lack of digital language resources like text labelled with corresponding sentiment classes. One reason for that is that manually labelling text data is time-consuming and expensive. Consequently, automatic and rapid processes are needed to reduce the manual effort as much as possible making the labelling process as efficient as possible. In this paper, we present and analyze an automatic language-independent sentiment labelling method that leverages information from sentiment-bearing emojis and words.  Our experiments are conducted with tweets in the languages English, Sepedi and Setswana from \textit{SAfriSenti}, a multilingual sentiment corpus for South African languages. We show that our sentiment labelling approach is able to label the English tweets with an accuracy of 66\%, the Sepedi tweets with 69\%, and the Setswana tweets with 63\%, so that on average only 34\% of the automatically generated labels remain to be corrected.}

\keywords{sentiment analysis, artificial intelligence, natural language processing, South African languages}

\begin{document}

\maketitleabstract

\section{Introduction}

Sentiment analysis helps analyze and extract information about polarity from textual feedback and opinions. Sentiment analysis draws attention in business environments~\cite{Rokade2019BusinessIA} and other areas, like medicine~\cite{Zucco:2018}, education~\cite{MabokelaTurgayRaborife2022,Rakhmanov+Schlippe:2022} and AI for Social Good~\cite{MabokelaSchlippe:2022b}.

Sentiment analysis for under-resourced language still is a skewed research area. 
Although, there are some considerable efforts in emerging African countries to develop resources for under-resourced
languages, some languages such as indigenous South African languages still suffer from a lack of datasets. One reason for that is that manually labelling text data is time-consuming and expensive. Consequently, automatic and rapid processes are needed to reduce the manual effort as much as possible making the labelling process as efficient as possible. In this paper, we present and analyze an automatic language-independent sentiment labelling algorithm that leverages information from sentiment-bearing emojis\footnote{Emojis are pictorial representations of emotions, ideas, or objects in electronic communication to add emotional context.} and words. We will evaluate our algorithm on a subset of our \textit{SAfriSenti} corpus \cite{MabokelaSchlippe2022,MabokelaSchlippe:2022b} with English, Sepedi and Setswana tweets. Sepedi is mainly spoken in the northern parts of South Africa by 4.7~million people and Setswana by 4.5~million people~\cite{statista:Africa}.

In the next section, we will describe related work. In section~3 we will present our language-independent algorithm for sentiment labelling. The experimental setup will be characterised in Section~4. In Section~5 we will summarise the results of our experiments. 
We will conclude our work in Section~6 and indicate possible future work. 

\vspace{-1.2pt}

\section{Related Work}

Previous studies investigated sentiment data collection strategies for under-resourced languages on Twitter \cite{paktwitter2010,Vosoughi2016}. The methods focus on labelling only two sentiment classes ---positive and negative.
Meanwhile other research work has explored strategies to label three sentiment classes in Twitter---positive, neutral, and negative  ---using human annotators \cite{pang2002thumbs,paktwitter2010,vilares2016cs,nakov2019semeval}. Despite the attempt to automate the data labelling process \cite{Kranjc2015ActiveLF}, the hand-crafted annotation is to date the most preferred method of data labelling in many natural language processing tasks \cite{Chakravarthi2020CorpusCF}. 
However, manual annotation presents challenges and it is deemed an expensive process. Notably, \cite{Jamatia2020,Gupta2021UnsupervisedSF} employed manually annotated tweets, while other studies focus on automated data labelling solutions \cite{Kranjc2015ActiveLF}. \cite{Vosoughi2016} investigated various pipelines to collect data on Twitter using distant supervised learning. In this approach, they use positive and negative emojis as indicators to annotate tweets. 

\cite{Go2009} explored distant supervision methods to label millions of tweets using positive and negative search terms (i.e.\ term queries) in the Twitter API and emojis to pre-classify the tweets.
\cite{vilares2016cs} investigated \textit{SentiStrength} scores to label an English-Spanish code-switching Twitter corpus. \textit{SentiStrength} is an online sentiment analysis system available for a few languages \cite{Thelwall2011SentimentIT}.

Compared to \cite{vilares2016cs,cliche2017bb_twtr,Jamatia2020}, we also investigate a distant supervised annotation method. However, we automatically build up lists with sentiment-bearing words after we have made use of emojis as indicators for the sentiment classes. This way we can label all tweets---first those tweets that contain sentiment-bearing emojis, and then the rest of the tweets based on the words in the tweets with the sentiment-bearing emojis.

\section{Sentiment Labelling with Distant Supervision}

\begin{figure}[h!]
  \begin{center}
  \includegraphics[width=1.0\linewidth]{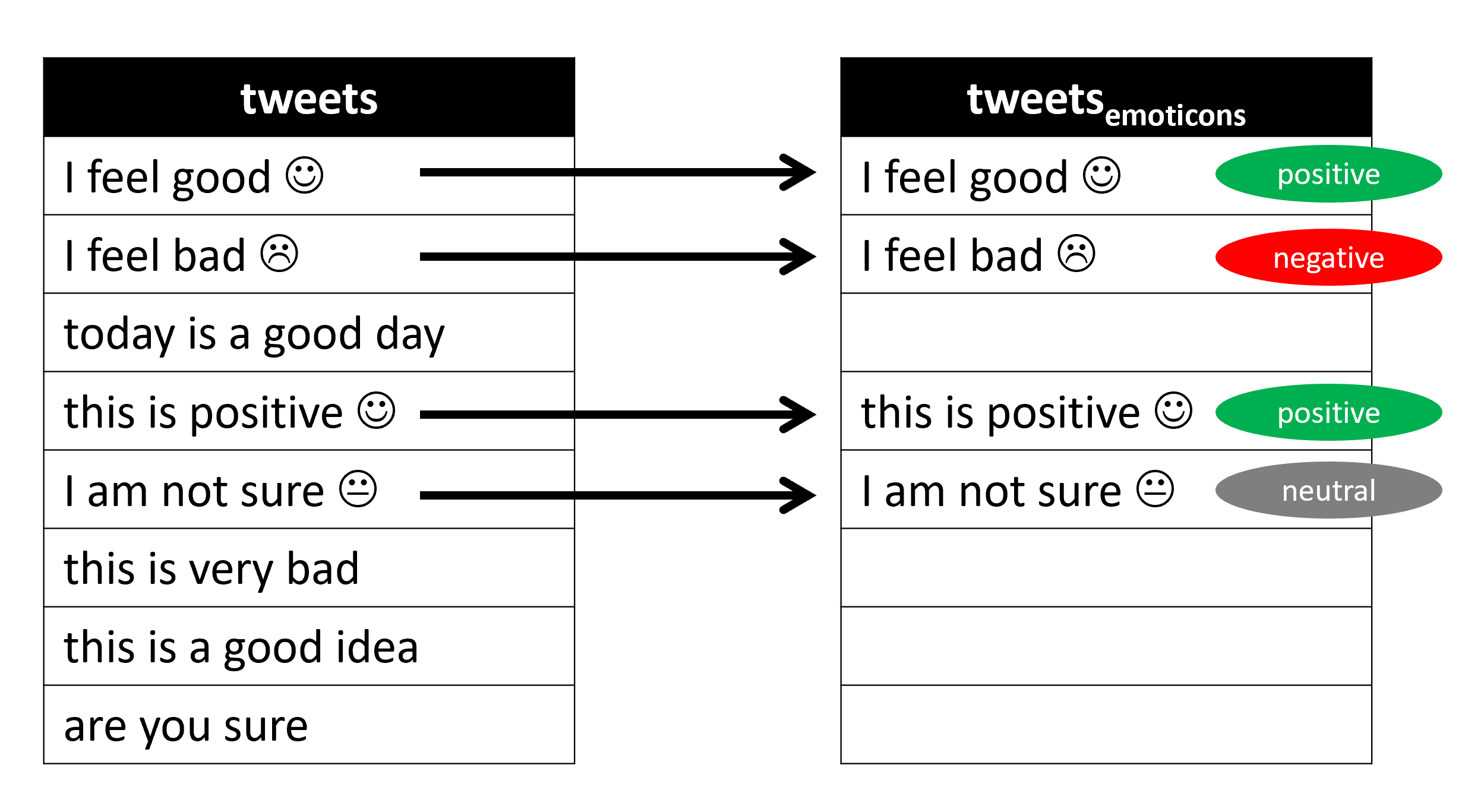}
  \vspace{-20.2pt}
  \caption{Classify tweets with sentiment-bearing emojis into the 3 classes ($step1_{emojis}$).}
  \label{fig:step1}
  \end{center}
\end{figure}

\begin{figure}[h!]
  \centering
  \includegraphics[width=1.0\linewidth]{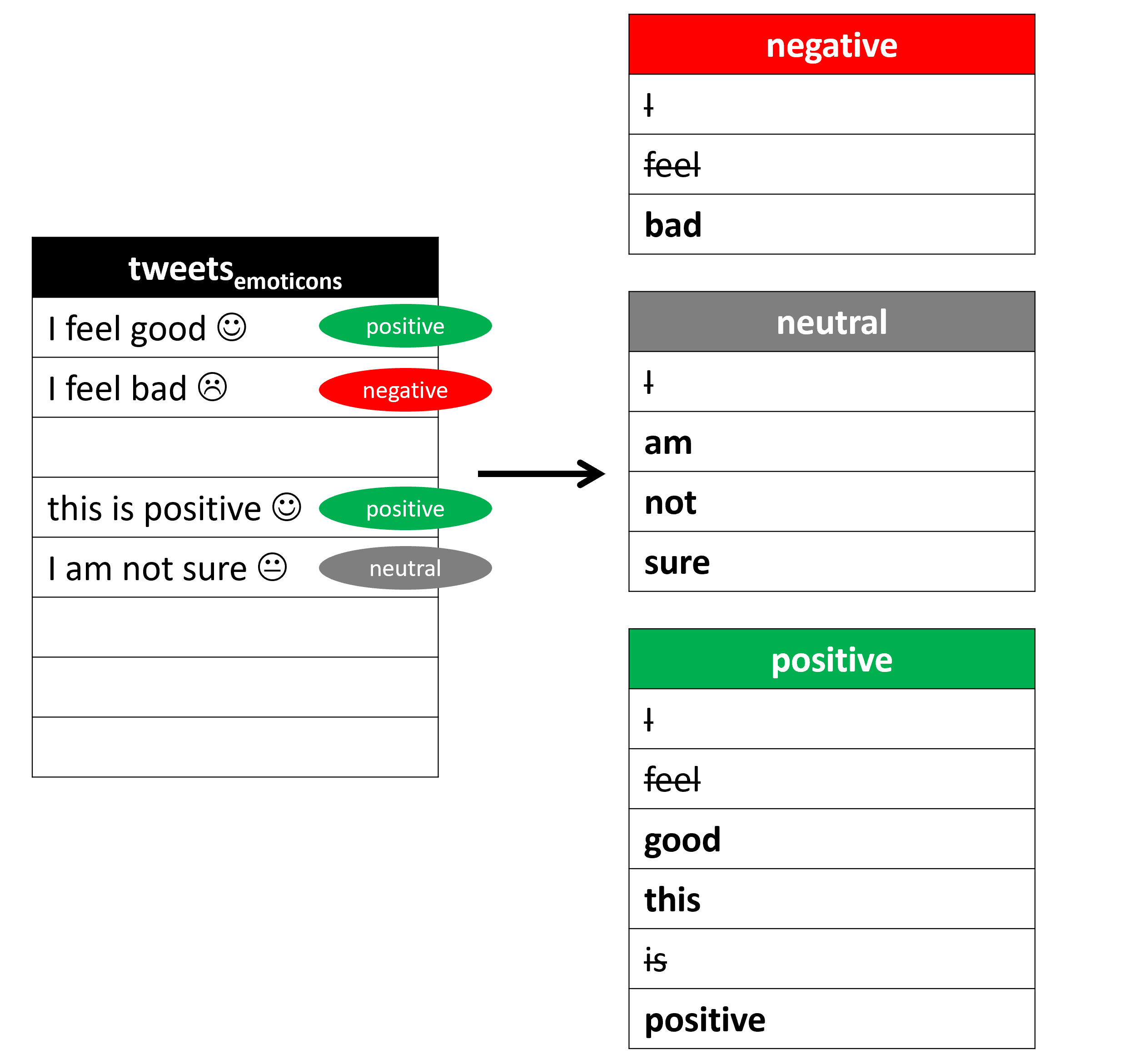}
  \vspace{-20.2pt}
  \caption{Create lists with sentiment-bearing words ($step2_{lists}$).}
  \label{fig:step2}
\end{figure}

\begin{figure}[h!]
  \centering
  \includegraphics[width=1.0\linewidth]{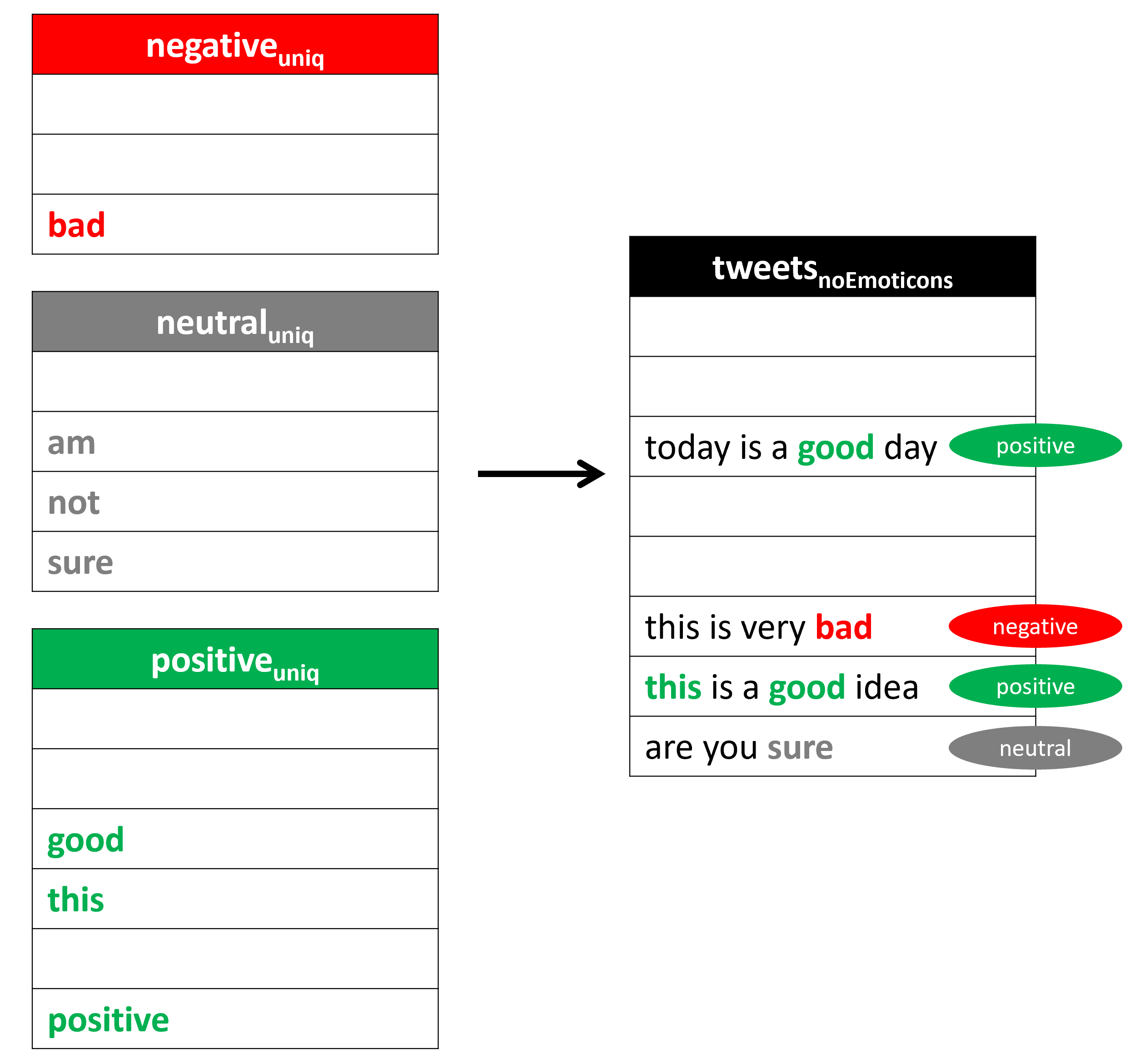}
  \vspace{-20.2pt}
  \caption{Sentiment-bearing words as indicators for remaining tweets' sentiment classes ($step3_{words}$).}
  \label{fig:step3}
\end{figure}

As illustrated in Figure~\ref{fig:step1}, \ref{fig:step2} and \ref{fig:step3}, we propose the following algorithm for sentiment labelling that leverages information from sentiment-bearing emojis and words: 

\begin{itemize}
\item $Step~1_{emojis}$: Classify tweets with sentiment-bearing emojis into the classes \textit{negative}, \textit{neutral} and \textit{positive} (Figure~\ref{fig:step1}). 
\item $Step~2_{lists}$: Create lists with sentiment-bearing words (Figure~\ref{fig:step2}): \begin{enumerate}
\item Collect all words from \textit{negative}, \textit{neutral} and \textit{positive}.
\item Then remove words that occur in one or both other lists.
\end{enumerate}
\item $Step~3_{words}$: Classify remaining tweets without sentiment-bearing emojis into the classes \textit{negative}, \textit{neutral} and \textit{positive} based on the highest word coverage with the lists of sentiment-bearing words (Figure~\ref{fig:step3}).

\end{itemize}


\vspace{-2.2pt}

\section{Experimental Setup}

In this section, we will describe the dataset for our experiments and how we used emojis as indicators for sentiments.

\subsection{SAfriSenti}

To evaluate our algorithm for sentiment labelling with comparable numbers of tweets in three languages, we applied it to monolingual 7,000~English tweets, 7,000~Sepedi tweets and~7,000 Setswana tweets from the \textit{SAfriSenti} corpus. \textit{SAfriSenti} is to date the largest sentiment dataset available for South African languages with 64.3\% of monolingual tweets in English, Sepedi and Setswana and 36.6\% of code-switched tweets between these languages~\cite{MabokelaSchlippe2022}. The monolingual tweets' distributions of the classes \textit{negative}, \textit{neutral}, \textit{positive} are demonstrated in Table~\ref{tab:English}, \ref{tab:Sepedi} and~\ref{tab:Setswana}.

\vspace{-1.2pt}

\begin{table}[!ht]
\centering
\footnotesize
\begin{tabular}{lrr}
\hline
Class       &   Number    & \%   \\
 \hline
positive       & 2,052    &   29.3      \\
negative       & 3,448       & 49.3       \\
neutral       & 1,500       & 21.4      \\
\hline
Total       & 7,000     &       \\
\hline
\end{tabular}
\caption{Distribution of English tweets.}
\label{tab:English}
\end{table}

\vspace{-1.2pt}

\begin{table}[!ht]
\centering
\footnotesize
\begin{tabular}{lrr}
\hline
Class       &   Number    & \%   \\
 \hline
positive       & 3,500       & 50.0        \\
negative       & 2,270     & 32.4       \\
neutral       & 1,230      & 17,6      \\
\hline
Total       & 7,000      &       \\
\hline
\end{tabular}
\caption{Distribution of Sepedi tweets.}
\label{tab:Sepedi}
\end{table}

\vspace{-1.2pt}

\begin{table}[!ht]
\centering
\footnotesize
\begin{tabular}{lrr}
\hline
Class       &   Number    & \%   \\
 \hline
positive       & 3,230     & 46.1        \\
negative       & 2,180       &31.1       \\
neutral       & 1,590      & 22.8      \\
\hline
Total       & 7,000      &       \\
\hline
\end{tabular}
\caption{Distribution of Setswana tweets.}
\label{tab:Setswana}
\end{table}

\vspace{-2pt}

\subsection{Emojis}

For our experiments, we defined 12 emojis as \textit{negative} indicators, 10 emojis as \textit{neutral} indicators, and 12~emojis as \textit{positive} indicators as listed in Table~\ref{emojis}, for which were sure that they would represent the corresponding sentiments well. Of course, our emoji list can be extended based on further information such as the Emoji Sentiment Ranking\footnote{\url{https://kt.ijs.si/data/Emoji_sentiment_ranking}} \cite{Kralj2015emojis}.

\begin{table}[h]
\centering
\footnotesize
\begin{tabular}{lrrr}
\hline
sentiment & \#emojis\\
\hline
negative &  12\\
neutral &  10\\
positive & 12\\
\hline
\end{tabular}
\caption{\label{emojis}
Emojis for \textit{negative}, \textit{neutral} and \textit{positive}.
}
\end{table}

\begin{figure}[h!]
  \centering
  \includegraphics[width=0.7\linewidth]{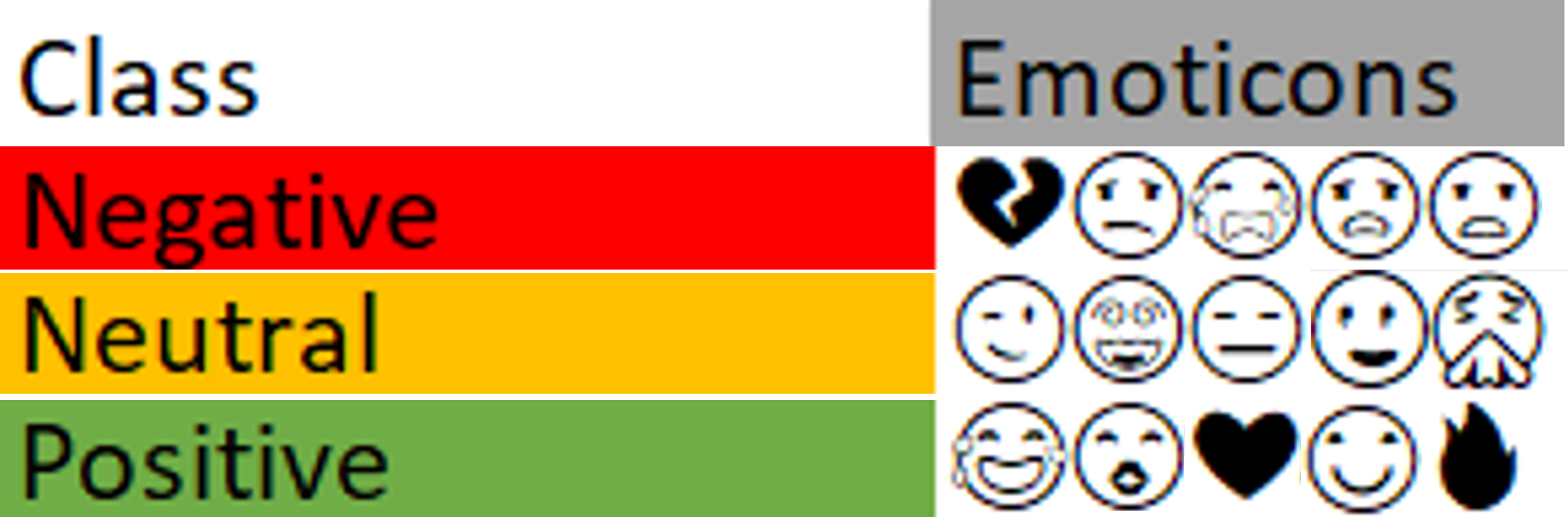}
    \vspace{-2.2pt}
  \caption{Examples of emojis}
  \label{fig:emojis}
\end{figure}

A subset of such emojis is illustrated in Figure~\ref{fig:emojis}. To be platform-independent, our algorithm finds and compares the emojis in Unicode. If a tweet contains multiple emojis representing different sentiments, the tweet is labelled with the sentiment class which has the most emojis in the tweet. The tweet is not labelled at this step if there is no majority.

\vspace{-5.2pt}

\vspace{-7.2pt}

\section{Experiments and Results}

We evaluated our algorithm for sentiment labelling on the English, Sepedi and Setswana tweets from \textit{SAfriSenti}. Table~\ref{percentages} demonstrates the absolute numbers and percentages of tweets with sentiment-bearing emojis in $step~1_{emojis}$ and the absolute numbers and percentages of the remaining tweets which were labelled using our sentiment-bearing words in $step~3_{words}$.

\begin{table}[h]
\centering
\footnotesize
\begin{tabular}{lrr}
\hline
language & \textbf{$step~1_{emojis}$} & \textbf{$step~3_{words}$}\\
\hline
English & 4,210 (60.1\%) & 2,790 (39.9\%) \\
Sepedi & 5,871 (83.9\%) & 1,129 (16.1\%) \\
Setswana & 3,249 (46.4\%) & 3,751 (53.6\%) \\
\hline
\end{tabular}
\caption{\label{percentages}
\#tweets and \%tweets for $step1$ and $step3$.
}
\end{table}

\begin{table}[h!]
\centering
\footnotesize
\begin{tabular}{lrrr}
\hline
language & \textbf{$step~1_{emojis}$} & \textbf{$step~3_{words}$} & $step~1$-to-3\\
\hline
English & 68.7\% & 63.5\% & 66.2\% \\
Sepedi & 69.5\% & 64.6\% & 68.7\% \\
Setswana & 66.1\% & 59.7\% & 62.7\%\\
\hline
\end{tabular}
\caption{\label{Accuracies}
Accuracies.
}
\end{table}

\begin{table}[h!]
\centering
\footnotesize
\begin{tabular}{lrrr}
\hline
language & \textbf{$step~1_{emojis}$} & \textbf{$step~3_{words}$} & $step~1$-to-3\\
\hline
English & 66.8\% & 62.2\% & 64.6\% \\
Sepedi & 68.2\% & 63.4\% & 67.9\% \\
Setswana & 65.2\% & 58.7\% & 61.5\% \\
\hline
\end{tabular}
\caption{\label{F-scores}
F-scores.
}
\end{table}

Table~\ref{Accuracies} and Table~\ref{F-scores} shows the accuracies and F-scores of automatically classifying tweets with sentiment-bearing emojis in $step~1_{emojis}$, the accuracies and F-scores of  automatically classifying the remaining tweets using our sentiment-bearing words in $step~3_{words}$ as well as the accuracies and F-scores of all 7k labeled tweets together ($step~1$-to-3). For computing the F-scores, we took the \textit{Macro $F_{1}$ Score Calculation}, i.e.\ the average of each class’s $F_{1}$ score.

The results of the two tables demonstrate that the quality of the automatic labeling in $step~1_{emojis}$ is better than that in $step~3_{words}$. This shows that in $step~2_{lists}$ the automatically created word lists are worse indicators than the emojis. The goal of our algorithm is to label a large part correctly so that the annotators no longer have to label all labels from scratch and thus minimize the workload. From the accuracies and the F-scores, we see that we were able to reduce the manual effort, but there is still room for improvement: With an accuracy of 66\%, the annotators would still have to change 34\% of the labelled tweets for English. Our algorithm performs best for Sepedi with 68\% accuracy, which would require 32\% changes. With Setswana, the accuracy is 63\%, which is why 37\% of the labels would have to be changed.

Both the quality of $step~1_{emojis}$ but also the quantity of tweets with sentiment-bearing emojis in the corpus, based on which the word lists are generated in $step~2_{lists}$, have an impact on the quality of the labeling in $step~3_{words}$: The more tweets with sentiment-bearing emojis, the higher the chance that qualified sentiment-bearing words remain in the lists after $step~2_{lists}$. We believe that Sepedi especially performs best since out of the 7k Sepedi tweets, 84\% contain sentiment-bearing emojis and thus more text can be used for the generation of the word lists than for English (60\%) and Setswana (47\%). Still, with an accuracy of 70\%, the challenge remains to figure out how to use emojis as a better indicator. 

Adding more features for pre-labeling such as a translation of the tweets or word lists translated from other languages could help. But then the algorithm would no longer be completely language-independent and would have to deal with information coming from outside the corpus. 

\vspace{-2.2pt}

\section{Conclusion and Future Work}
In this paper, we have presented a language-independent algorithm for sentiment labelling. Our algorithm uses sentiment-bearing emojis as initial features to build lists with sentiment-bearing words. Since our approach is only based on frequencies, no training of a machine learning system is required. This ways we completely avoid any manual or higher computational effort.

Our analyses on the under-resourced languages Sepedi and Setswana plus English demonstrated that accuracies between 63\% and 69\% are possible using our distant supervision approach. This significantly reduces the manual effort to label tweets with sentiment classes since the human annotators need to change between 31\%--37\% of the tweets that have been pre-labelled with our algorithm instead of adding all labels from scratch. After our proof of concept with three languages, it can be assumed that our approach works for other languages as well, since people often use emojis in their posts---no matter what language their posts are.

Consequently, it is interesting to apply our approach to more languages and to experiment with cross-lingual features. Future work may also include investigating whether it is helpful to label a tweet as \textit{neutral} if it contains a comparable number of \textit{positive} and \textit{negative} emojis. Our current algorithm does not assign a label to the tweet in this case. Furthermore, we plan to combine our approach with active learning, i.e.\ an iterative process where annotators manually classify tweets which are then used to re-train machine learning systems for classification. 
Additionally, our goal is to create a multilingual natural language processing model and investigate the synergy effects across languages in sentiment analysis. \cite{Makgatho_2021}'s word embeddings could be a good basis for such a model.


\bibliographystyle{ltc23}
\bibliography{xample23.bib} 

\end{document}